\documentclass[letterpaper, 10 pt, conference]{ieeeconf}  

\IEEEoverridecommandlockouts                              

\usepackage{graphics} 
\graphicspath{ {./Figures/} }
\usepackage{epsfig} 
\usepackage{amsmath} 
\usepackage{amssymb}  
\usepackage{hyperref}
\usepackage[capitalise]{cleveref}

\title{\LARGE \bf AutoCone: An OmniDirectional Robot for Lane-Level Cone Placement}

\author{Jacob Hartzer and Srikanth Saripalli$^{1}$
\thanks{$^{1}$Jacob Hartzer and Srikanth Saripalli are with the Department of Mechanical Engineering, Texas A\&M University, College Station, Texas, USA {\tt\small jmhartzer@tamu.edu}, {\tt\small ssaripalli@tamu.edu}}%
}

\usepackage{tikz}
\usepackage{textcomp}
\usepackage{lipsum}

\newcommand\copyrighttext{%
  \footnotesize \textcopyright 2020 IEEE. Personal use of this material is permitted.
  Permission from IEEE must be obtained for all other uses, in any current or future
  media, including reprinting/republishing this material for advertising or promotional
  purposes, creating new collective works, for resale or redistribution to servers or
  lists, or reuse of any copyrighted component of this work in other works.
  DOI: \href{https://doi.org/10.1109/IV47402.2020.9304683}{10.1109/IV47402.2020.9304683}}
\newcommand\copyrightnotice{%
\begin{tikzpicture}[remember picture,overlay]
\node[anchor=south,yshift=10pt] at (current page.south) {\fbox{\parbox{\dimexpr\textwidth-\fboxsep-\fboxrule\relax}{\copyrighttext}}};
\end{tikzpicture}%
}

\begin{document}

\maketitle
\copyrightnotice
\thispagestyle{empty}
\pagestyle{empty}

\begin{abstract}

    This paper summarizes the progress in developing a rugged, low-cost, automated ground cone robot network capable of traffic delineation at lane-level precision. A holonomic omnidirectional base with a traffic delineator was developed to control for initialization errors. RTK GPS was utilized to reduce minimum position error to 2 centimeters. Due to recent developments, the cost of the platform is now less than \$1,600. To minimize the effects of GPS-denied environments, wheel encoders and an Extended Kalman Filter were implemented to maintain lane-level accuracy during operation and a maximum error of 1.97 meters through 50 meters with little to no GPS signal. Future work includes increasing the operational speed of the platforms, incorporating lanelet information for path planning, and cross-platform estimation.

\end{abstract}

\section{INTRODUCTION}

The North Texas Tollway Authority (NTTA) is an organization that runs and maintains toll roads, bridges, and tunnels in North Texas. They are responsible for collecting tolls and to use those tolls to fund projects and services connected to the infrastructure and use of these roads. This includes servicing the roadway itself, mobile repairs or service, and the monitoring of roads for safety and traffic conditions. This allows the NTTA to respond and adapt to changing conditions throughout the day and to better serve those who use their roads.

\begin{figure}
    \centering
    \includegraphics[width = \linewidth]{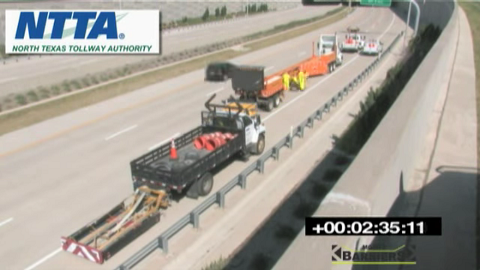}
    \caption{Attenuator Truck Setup During Highway Work}
    \label{fig:AttenuatorTruck}
\end{figure}

Due to the work performed by NTTA, it must be possible to access and safely cordon portions of the road to work or provide assistance to drivers. To do this, NTTA has two attenuator trucks capable of withstanding a rear or side impact, which protect road workers while not constricting access to the road itself. Additionally, there are numerous response trucks used in service but that are also used to block off a lane from the rear. Shown in \cref{fig:AttenuatorTruck}, the service truck in addition to the attenuator truck can be used to block off multiple lanes while still allowing access to work within the lane \cite{NTTA}.

These attenuator trucks and service trucks are vital to the safety of the NTTA workers due to the unpredictability of drivers at all times of day. However, there have been issues with the number of vehicles that have collided with these trucks. Despite working as designed to protect the workers, the attenuator trucks have long down times after collision and a high cost to repair. At times, both trucks have been out of service, preventing the workers from being able to safely service the road or aid drivers. And while the response trucks can also be used to block traffic and protect workers, collisions with these trucks have the same issue of large costs for repairs.

NTTA has found that placing cones in a wedge shape behind the vehicles drastically reduces the number of collisions with the trucks by giving drivers additional warning of upcoming road work. This is especially effective around corners and roads with high speed limits. The trade off is the placement of cones requires the workers to manually drop off and pick up the cones up to 120 feet away from the protection of the service and attenuator trucks. Due to the high risk of personal injury to workers, this was deemed unacceptable. Therefore, the NTTA desired a platform removing the need for highway workers to manually place cones in the road. This system would need to be low-cost due to high risk of being struck by vehicles, fully automated and rugged for operation on highways, and precise in remaining within lanes and cone placement.

This paper outlines the progress towards the creation of a low-cost omnidirectional robot that has lane-level cone placement.

\section{PROBLEM DEFINITION}

The goal of this project is to develop a robotic platform capable of automatically placing cones in a defined wedge shape behind the work vehicle within the starting lane.

The system shall:
\begin{itemize}
    \item Place three cones in 40 foot increments in a wedge
    \item Begin the wedge 80 feet from the end of the vehicle
    \item Operate on highway surfaces unaffected by small debris
    \item Remain within the lane despite road curvature
    \item Not rely on magnetic or road-embedded sensors
    \item Have a speed greater than 0.3 m/s
    \item Cost less than \$1,500 per cone unit
    \item Be easy to use and require little training
\end{itemize}

\section{PLATFORM DESIGN SELECTION}

The given requirements and constraints were evaluated to determine various design decisions. Cost was a major constraint, and was considered in design and throughout the development of the project.

Under standard operation, it is not reasonable to assume the operators of the cones will be able to accurately or consistently place the cones on the roadway. With this  high uncertainly in initial state, it was desirable to have a holonomic robot capable of counteracting any placement error from the operator without requiring the platform to drive outside of the starting lane. This also has benefits when considering additional sensors, such as proximity sensors, that could cause discrete changes in the path of the robot.

\begin{figure}
    \centering
    \includegraphics[width= 0.75\linewidth]{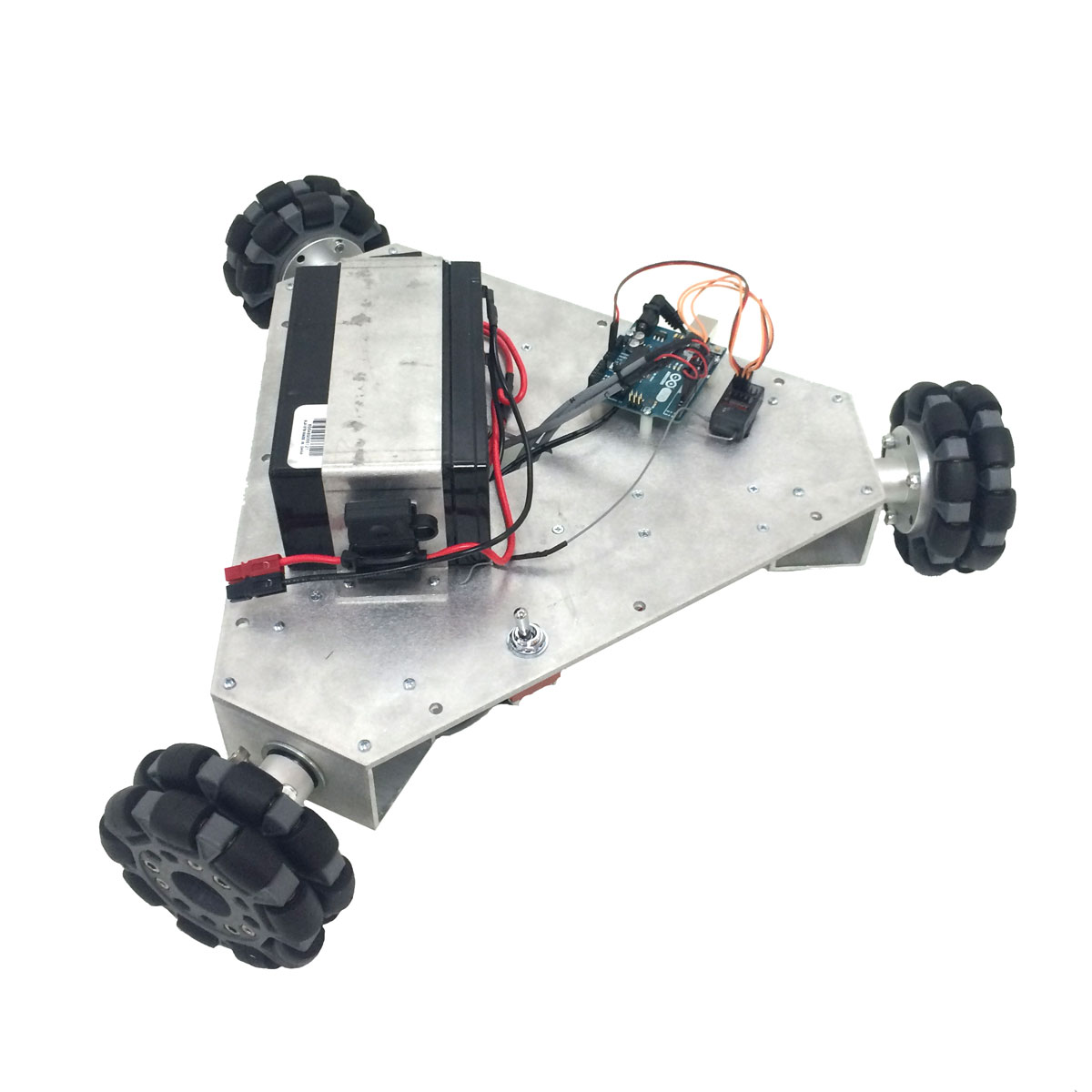}
    \caption{Omnidirectional Base from SuperDroid Robots}
    \label{fig:OmniBase}
\end{figure}

To make the robot holonomic, either four mecanum wheels or three omni wheels were to be used. When considering cost, an omni-wheeled system required one fewer motor, and therefore was selected for preliminary testing. Additionally, in order to maintain high position accuracy, motors with encoders were selected. With these considerations, the selected platform was the \textit{Programmable Tri-Wheel Vectoring Robot Kit}, available from SuperDroid Robots, shown in \cref{fig:OmniBase}, with IG32P motors and mounted encoders \cite{SuperDroid}. The platform was then modified to support two GPS antenna and an orange traffic delineator.

In order to control the platform, the system was fitted with a raspberry pi and two quadrature encoder counters to interface with the motor controllers and interpret sensor data. Beyond being relatively expensive, the raspberry pi 3B+ was capable of meeting performance needs for the project. The use of RTK GPS was an unexpected result of the release of the ZED F9P module from U-blox. This low-cost module brought high precision GPS positioning to the project, and is valuable in the navigation and control of the platform while only increasing the total platform cost to under \$1,600.

Cameras and LiDAR were also considered for the platform. Cameras are inexpensive, but due to the platform being so low to the ground, and the unreliability of cameras during sunrise and sunset, it was decided to progress without using cameras. Additionally, LiDAR sensors were considered, but not used. A sensor that could provide sufficiently accurate and dense data would be cost-prohibitive for this project, and the sensors on board the platform must be rugged to withstand vibrations from road use and transport.

\section{SYSTEM ARCHITECTURE DESIGN}

\subsection{Data Management}

\begin{figure}
    \centering
    \includegraphics[width = \linewidth]{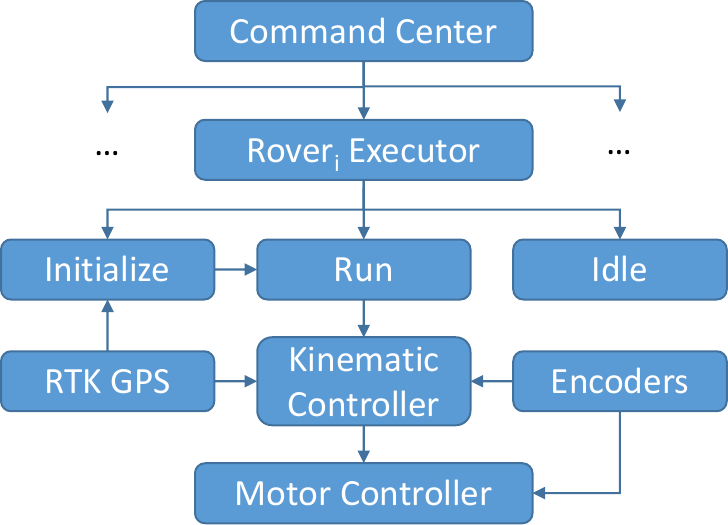}
    \caption{Information flow during platform operation}
    \label{fig:ConeFlow}
\end{figure}

In order to control these multiple robotic systems, it was important to consider the architecture between these different devices and how control and information were managed between them. The core of the architecture was formulated with the understanding that the cones will be deployed, recalled, and, if necessary, controlled from the main work truck. Because of this, the flow of high level commands should come from a main command center on the truck. The work trucks are equipped with low-power tablets regularly used by the operators which will serve as the command center. The data flow is outlined in \cref{fig:ConeFlow}.

From this point, the cones will use on-board computing to interpret the centralized signals as well as use their own internal state managers. The top-level executors will allow the cone to switch between an initializing state, where initial position is accurately estimated, and a running state, where the cone will follow a generated path to the final destination and back. This running state will accept deployment commands from the central command and position commands from its own executor. While running, the cone executor can receive status signals from a lower level process that determines the health of the cone's position estimation or the current path. This is calculated through GPS error estimates, but can be bolstered with proximity sensors, or computer vision. While running, the on board kinematic controller uses path and state information to determine desired local velocity, which is transformed into individual motor speeds using the kinematics of the tri-wheeled omni-robot from \cite{Baede}.

The individual motor speeds are communicated to two RoboteQ SDC2130 motor controllers. Using the quadrature encoder counters for feedback, PI velocity control is utilized. The velocity commands are given to the motor controllers through a serial connection using the roboteq package \cite{RoboteqIndigo}. An additional package called roboteq\_python was used to retrieve ROS messages for velocity commands and translate them into commands to the roboteq package \cite{RoboteqPython}.

With these considerations, incorporating additional states, sensing data, or control algorithms will be simple, due to the modularity of the code and architecture. New technologies are readily testable, and the barrier to pivot design choices will remain low.

\subsection{Real-Time Kinematic GPS}
\begin{figure}
    \centering
    \includegraphics{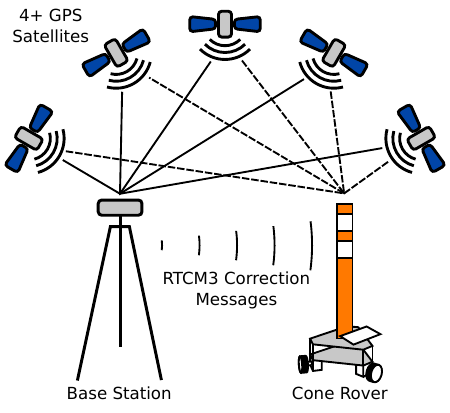}
    \caption{Real Time Kinematic GPS System}
    \label{fig:RTKGPS}
\end{figure}

To achieve high-precision GPS position estimates, the technique of real-time kinematic (RTK) GPS was utilized. This technique utilizes a immobile base station to solve the integer ambiguity search problem to determine the measured carrier phase. These carrier phase measurements are then sent to the rover as correction data shown in \cref{fig:RTKGPS}. The rover then uses the number of carrier cycles and multiplies by the wavelength, which is approximately 19 cm for the L1 signal. This technique results in accuracy up to 1 cm \cite{NovAtel}. The selected module to implement this technique is the ZED-F9H which has a baseline accuracy of 1 cm and a heading accuracy of 0.8 deg with a baseline of 0.25 m \cite{ZED}. Two ZED-F9P modules were implemented with differential heading using simpleRTK2B and simpleRTK2Blite boards from Ardusimple \cite{ArduSimple}.

Generally, according to surveying standards, temporary base station correction data is accurate within 10 km of the base station itself, creating a need to have closely available correction data on the highway \cite{NJGPS}. In order to ensure the base station is always within range of the rover, base stations will be part of the operator vehicle and command center computer or added to existing highway stations that are located every two miles.

\section{Kinematics}

Because the platform is holonomic, it can smoothly transition between any state. Position and heading are independent, which allows for a relatively simple kinematic controller.

\begin{figure}
    \centering
    \includegraphics[width = \linewidth]{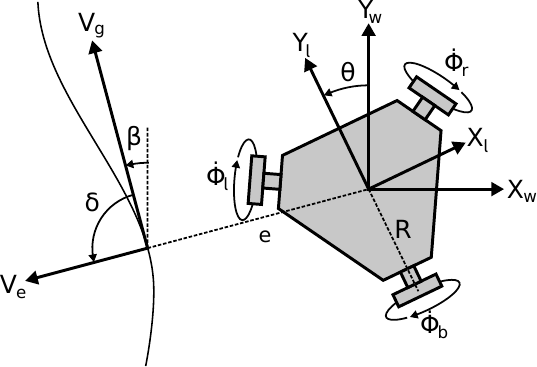}
    \caption{Kinematic Controller Model}
    \label{fig:Kinematics}
\end{figure}

The model used for the kinematic controller is shown in \cref{fig:Kinematics}. The command velocities in world coordinates are defined as the sum of position error velocity and the goal velocity from the nearest point on the path. This gives the following equation

\begin{equation}
    \begin{bmatrix}
        \dot{x}_w \\
        \dot{y}_w \\
        \dot{\theta}_w
    \end{bmatrix} =
    K\cdot e
    \begin{bmatrix}
        \sin{(\beta + \delta)} \\
        \cos{(\beta)}          \\
        0
    \end{bmatrix} +
    \begin{bmatrix}
        \dot{x}_g \\
        \dot{y}_g \\
        \dot{\theta}_g
    \end{bmatrix}
\end{equation}

where K is the error gain, $\beta$ is the path angle from the world Y axis, and $\delta$ gives the direction of the position error.

\subsection{Inverse Kinematics}
In order to utilize these commands, the world command velocities are then rotated to the local frame using the following equation

\begin{equation}
    \begin{bmatrix}
        \dot{x}_l \\
        \dot{y}_l \\
        \dot{\theta}_l
    \end{bmatrix}=
    \begin{bmatrix}
        \cos{\theta}  & \sin{\theta} & 0 \\
        -\sin{\theta} & \cos{\theta} & 0 \\
        0             & 0            & 1
    \end{bmatrix}
    \begin{bmatrix}
        \dot{x}_w \\
        \dot{y}_w \\
        \dot{\theta}_w
    \end{bmatrix}
\end{equation}

where $\theta$ is the angle of the local platform frame to the world frame.

Finally, the resultant local velocity command is scaled to the magnitude of the desired speed parameter, and transformed to the local wheel rotational velocities,

\begin{equation}
    \begin{bmatrix}
        \dot{\Phi}_l \\
        \dot{\Phi}_b \\
        \dot{\Phi}_r
    \end{bmatrix} =
    \frac{1}{r}
    \begin{bmatrix}
        -\frac{1}{2} & -\frac{\sqrt{3}}{2} & R \\
        1            & 0                   & R \\
        -\frac{1}{2} & \frac{\sqrt{3}}{2}  & R
    \end{bmatrix}
    \begin{bmatrix}
        \dot{x}_l \\
        \dot{y}_l \\
        \dot{\theta}_l
    \end{bmatrix}
\end{equation}

where $r$ is the radius of the omnidirectional wheels, $R$ is the radius from the center of gravity to the wheel point of contact, and $\Phi$ is the angular velocity command to each motor: left, back, and right respectively.

\subsection{Forward Kinematics}

In order to predict the motion of the platform, the forward kinematic equations are used. The transformation from motor angular velocities to local velocity is given by

\begin{equation}
    \begin{bmatrix}
        \dot{x}_l \\
        \dot{y}_l \\
        \dot{\theta}_l
    \end{bmatrix} =
    \frac{r}{3}
    \begin{bmatrix}
        -1          & 2           & -1          \\
        -\sqrt{3}   & 0           & \sqrt{3}    \\
        \frac{1}{R} & \frac{1}{R} & \frac{1}{R}
    \end{bmatrix}
    \begin{bmatrix}
        \dot{\Phi}_l \\
        \dot{\Phi}_b \\
        \dot{\Phi}_r
    \end{bmatrix}
\end{equation}

where the constants are defined in the previous section. These local velocities are transformed to world coordinates is given by the following rotation matrix.

\begin{equation}
    \begin{bmatrix}
        \dot{x}_l \\
        \dot{y}_l \\
        \dot{\theta}_l
    \end{bmatrix}=
    \begin{bmatrix}
        \cos{\theta} & -\sin{\theta} & 0 \\
        \sin{\theta} & \cos{\theta}  & 0 \\
        0            & 0             & 1
    \end{bmatrix}
    \begin{bmatrix}
        \dot{x}_w \\
        \dot{y}_w \\
        \dot{\theta}_w
    \end{bmatrix}
\end{equation}

\section{Extended Kalman Filter}

While the RTK-GPS normally gives an extremely accurate position and heading measurement at a relatively high rate, the platform will be required to operate in situations that deprive the antennae of GPS signal. In these GPS-denied environments, the platform should still be able to localize using wheel encoders. To combine these sensor data using the nonlinear state transition function, an extended Kalman filter (EKF) is used.

The \textit{a priori} state estimate $\hat{\boldsymbol{x}}$ is generated using the previously defined forward kinematics $f(\Phi)$ with the wheel encoder measurements.
\begin{equation}
    \hat{\boldsymbol{x}}_{k|k-1} = f(\dot{\boldsymbol{\Phi}}_{k-1})
\end{equation}
where the state is the North and East position (measured in meters from a set home position), heading angle with respect to North, and their derivatives.

The propagation of the \textit{a priori} covariance $\boldsymbol{P}$ is computed using the state transition matrix $\boldsymbol{F}$ and the process noise matrix $\boldsymbol{Q}$. The process noise covariance, which aids in predicting accuracy of state prediction, was estimated by running the platform under open loop control. The data was aggregated to produce an error covariance matrix which combines random walk and wheel slip errors.
\begin{equation}
    \boldsymbol{P}_{k|k-1} = \boldsymbol{F}_k \boldsymbol{P}_{k-1|k-1} \boldsymbol{F}^T_k + \boldsymbol{Q}_k
\end{equation}

The \textit{a posteriori} residual is generated using updates from RTK GPS, whose measurement function is given by $h(\hat{\boldsymbol{x}})$.
\begin{equation}
    \Bar{\boldsymbol{y}}_k = \boldsymbol{z}_k - h(\hat{\boldsymbol{x}}_{k|k-1})
\end{equation}
The residual covariance $\boldsymbol{S}$ is calculated using the linearized state observation matrix $\boldsymbol{H}$ and the observation noise matrix $\boldsymbol{R}$. Observation noise was calculated from the dilution of precision which is estimated from the ZED F9P RTK GPS module.
\begin{equation}
    \boldsymbol{S}_k = \boldsymbol{H}_k \boldsymbol{P}_{k|k-1} \boldsymbol{H}^T_k + \boldsymbol{R}_k
\end{equation}
The near-optimal Kalman gain $\boldsymbol{K}$ is given by
\begin{equation}
    \boldsymbol{K}_k = \boldsymbol{P}_{k|k-1} \boldsymbol{H}^T_k \boldsymbol{S}^{-1}_k
\end{equation}
which can be used to produce the updated state estimate
\begin{equation}
    \hat{\boldsymbol{x}}_{k|k} = \hat{\boldsymbol{x}}_{k|k-1} + \boldsymbol{K}_k \Tilde{\boldsymbol{y}}_k
\end{equation}
and the updated state covariance
\begin{equation}
    \boldsymbol{P}_{k|k} = (\boldsymbol{I} - \boldsymbol{K}_k \boldsymbol{H}_k ) \boldsymbol{P}_{k|k-1}
\end{equation}
Given this structure for the Kalman filter, the system will only update when a RTK GPS measurement is received and otherwise rely on dead reckoning based on wheel encoders.

\section{RESULTS}

\subsection{Dead Reckoning}

Using closed loop velocity control from the motor controllers, it was possible to relatively accurately control the position of the base using open-loop velocity commands.

When tested on smooth flat concrete, this process was found to have a horizontal position error of 1 foot at 80 feet travelled or 0.83\%. This is a key dimension as the robot cannot drift outside the lane else it will collide with oncoming traffic and fail to properly make the wedge shape to control the flow of vehicles. Because of this low value for a relatively inexpensive control system, further testing was done on actual highway surfaces.

On actual highway surfaces, it was found that errors in initial positioning and the uneven surface caused drift. Any slope of the road and the hard plastic wheels interacting with road debris caused the base to lose its initial heading. This drift accumulated to a horizontal error of 8 feet at 120 feet, or 6.67\%. This level of error is unacceptable in order to keep the system within a lane and is highly dependent on the skill of the operator. However, these results did increase the confidence of the accuracy of the motor encoders when combined with some other position feedback sensor.

\subsection{GPS Proportional Control Results}

Using an RTK GPS and base station allowed for much higher global accuracy of the cone. When testing on the highway, the RMS position error of the ZED-F9P  was 2 cm. This allowed for much higher repeatability and greatly reduced the drift that gravel or other rough surfaces could have on the heading.

Given this high-precision position and heading information, it was possible to create a proportional controller that gives linear and angular velocity commands based on the system error from a desired state. Control over the system was achieved by changing the heading and position desired state.

\begin{figure}
    \centering
    \includegraphics[width = \linewidth]{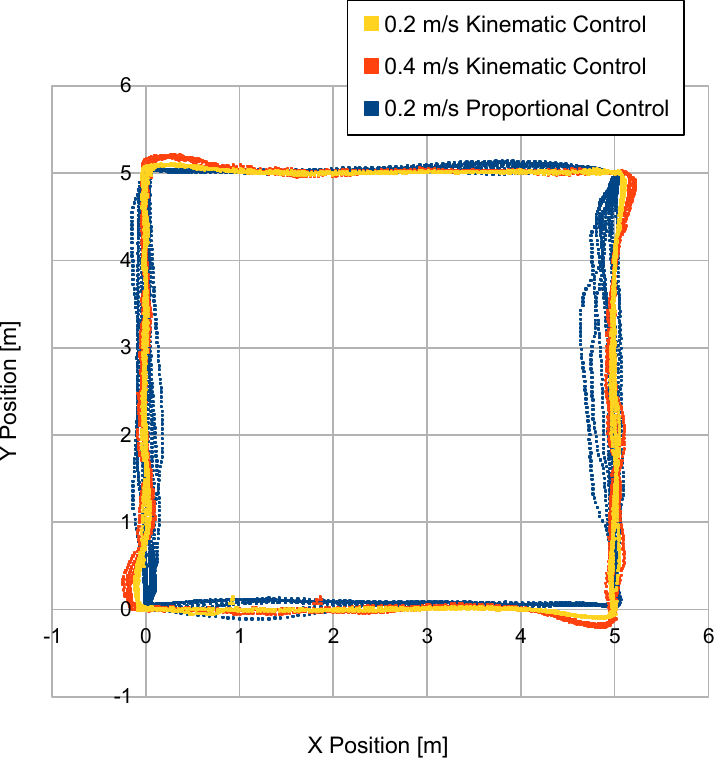}
    \caption{Square Cycles with Various Control Schema}
    \label{fig:SquareTest}
\end{figure}

To test the repeatability of the proportional controller, the platform was given consecutive corners of a 5 meter box. The results are shown as the Proportional Control plot in \cref{fig:SquareTest}. Despite 10 consecutive box movements for a total of 200 meters of motion, the proportional controller consistently placed the platform in the desired location with little final position error. However, drift and initial heading errors when beginning motion caused intermittent path errors of no more than 0.5 meters. By not considering path error, the proportional controller is much more likely to allow the platform to drift away from straight line motion.

\subsection{GPS-Denied Environments}
As many highways pass under roads, it was important to evaluate accuracy and availability of position and heading data in GPS-deprived environments. Therefore, exploratory tests in depriving the ZED-F9P of signal were conducted.

\begin{figure}
    \centering
    \includegraphics[width = 0.9\linewidth]{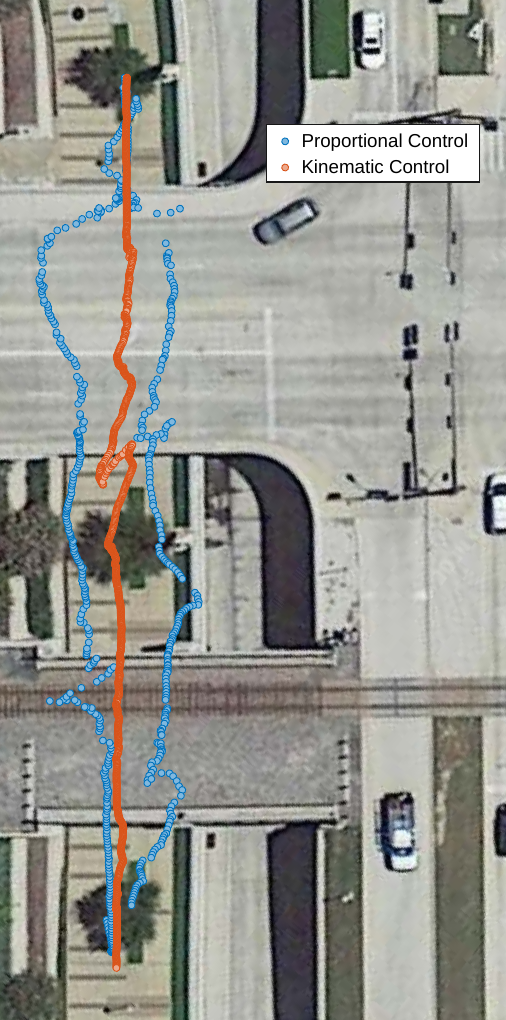}
    \caption{GPS-Denied Environment Testing}
    \label{fig:BridgeTest}
\end{figure}

The RTK GPS system was setup near a multi lane pedestrian and vehicle underpass where antennae would be obstructed for more than 50 meters. The platform was moved in a straight line along the road while collecting the GPS data. This data would represent the position estimate that would have been used by the platform with the proportional controller and is shown in the Proportional Control plot of \cref{fig:BridgeTest}. Going under this bridge caused a loss of high-precision GPS position and differential heading. The maximum error was 7.72 meters from the line of motion, with a RMS error of 3.51 meters. Interestingly, there was also a peak in error when moving under the set of train tracks. Once past the bridge, the system was able to quickly regain signal. However, the error far exceeded the width of a typical lane line and is therefore not acceptable. Because of this test, the EKF and kinematic controller were developed for the platform.

\subsection{ EKF and Kinematic Control Results}

Using the EKF and kinematic controller reduced the path error resulting from defects on the ground or initial heading errors while moving. Shown with the Kinematic Control plots of \cref{fig:SquareTest}, the kinematic controller had much lower path error, at the expense of overshoot errors at the corners. This is caused by the naive approach to the speed controller that has a constant goal velocity. As such, the platform moves at the goal velocity into the corner without deceleration, causing overshoot. Using a slower goal speed resulted in lower overshoot errors. However, using a separate speed controller that analyzes path curvature will allow the platform to have no overshoot and achieve near-perfect path following that is possible with the omnidirectional wheels.

Additionally, the GPS-denied environment test was repeated using a one-way path that extended under the bridge. The resulting state estimation from the EKF is shown in the Kinematic Control plot of \cref{fig:BridgeTest}. There was a maximum path error of 1.97 meters and a RMS error of 0.46 meters. The main deviation occurred when a GPS fix was regained but had high error due to occlusion from the road. Even at this point, the platform had error less than half the width of a lane. Therefore, the platform should be capable of navigating a path down the center of a lane without deviating outside, even in GPS-denied environments.

\section{FUTURE WORK}

Having shown the cone platform is capable of accurately following various paths, it must also be able to plan these paths. To achieve this, lanelet information will be incorporated to determine curvature and road location, which can be used to guide the cones along the center of the lane and correct for uncertainty of the road direction.

\begin{figure}
    \centering
    \includegraphics[width = 0.75\linewidth]{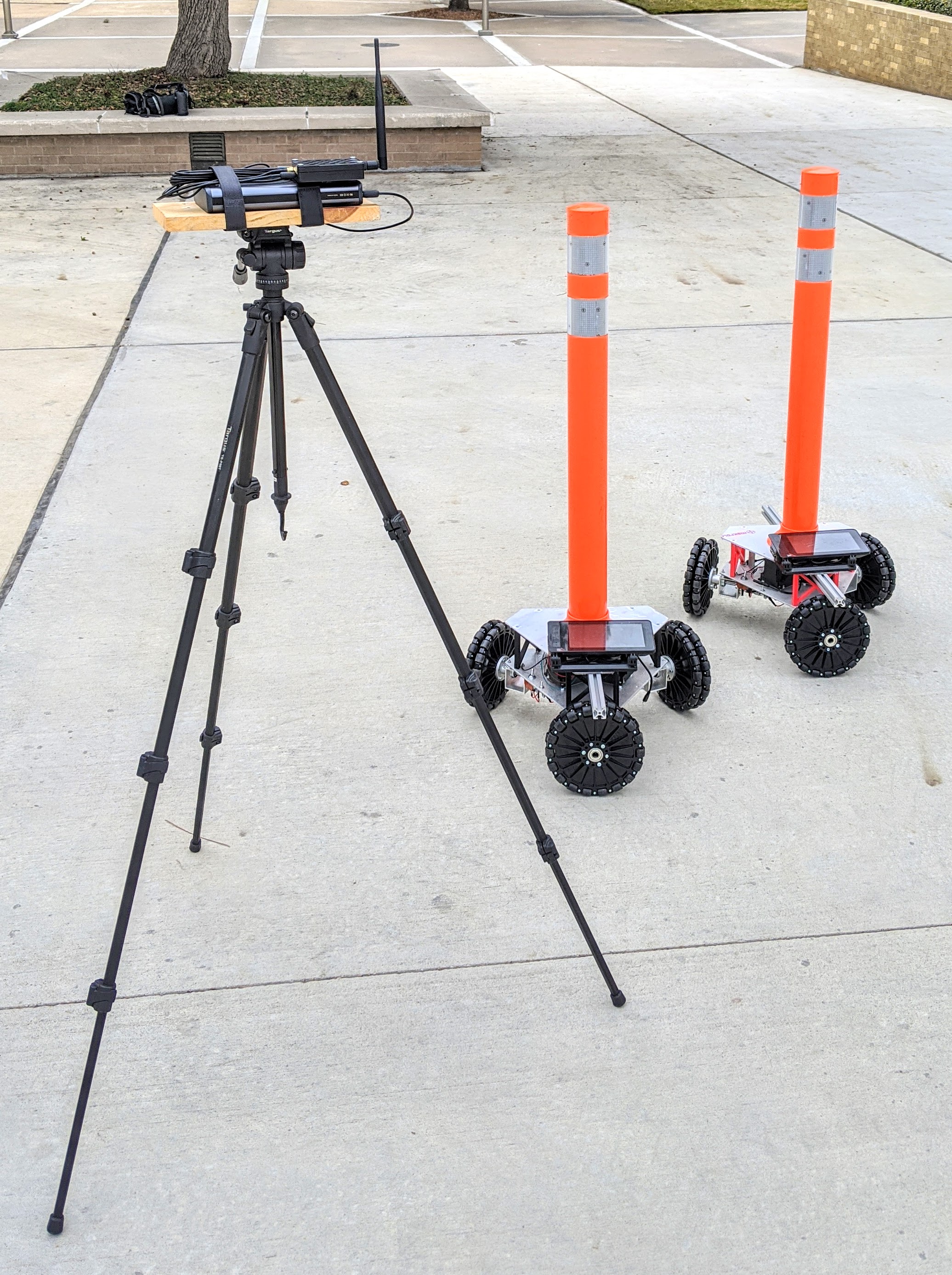}
    \caption{RTK Base Station and Two Platforms with Larger Omni Wheels}
    \label{fig:Setup}
\end{figure}

Additionally, feedback has shown the current operating speed of 0.2 m/s is too slow to keep workers from wanting to place the cones themselves. As such, larger diameter omni wheels shown in \cref{fig:Setup} are being prepared for testing to increase the operational speed of the platform without sacrificing accuracy or stability.

Finally, multiple platform coordination will be tested. Through networking, a more robust Kalman filter can be created with delayed state measurements across the platforms. This will require refinement of the platform level executor and represent a large step toward the goal of a prototype deployment of the cones for NTTA.

\section{CONCLUSIONS}

This project has developed a proof of concept omnidirectional cone platform that is able to maintain lane level accuracy while costing less than \$1,600. A holonomic omnidirectional platform was selected to correct for initial state errors and ensure lane boundaries are kept. The platform and sensors were also selected in order to be rugged, simple, and low cost. The system architecture was designed to allow control of multiple robots with individual state machines as well as information flow between robots. Additionally, the system architecture allowed the code and drivers to remain modular and adaptable. The platform can operate on highway surfaces while maintaining a dead-reckoning accuracy of less than 7\%, despite drift and initialization errors. RTK GPS was utilized for high-precision position accuracy of less than 2 cm while also providing heading from differential GPS. The implementation of a kinematic controller and Kalman filter allowed for low path error of the platform and successful operation in GPS-denied environments. Despite travelling over 50 meters in this environment, the platform was able to remain within lane-with, with a mazimum error of 1.97 meters. As the project progresses, higher operational speeds will be tested to ensure the platform is not slower than human operators placing cones. Additionally, lanelet information will be incorporated into path planning, to ensure the operation is automatic. Finally, multiple platform networking will be added to allow for high level remote control and higher order state estimation between platforms.






\section*{DISCLAIMER}

The contents of this report reflect the views of the authors, who are responsible for the facts and the accuracy of the information presented herein. This document is disseminated in the interest of information exchange. The report is funded, partially or entirely, by a grant from the North Texas Tollway Authority. However, NTTA assumes no liability for the contents or use thereof.

\section*{ACKNOWLEDGMENT}

Support for this research was provided in part by a grant from the North Texas Tollway Authority.


\bibliographystyle{IEEEtran}
\bibliography{IEEEabrv,references}

\end{document}